\documentclass[letterpaper]{article} % DO NOT CHANGE THIS
\usepackage{aaai2026}  % DO NOT CHANGE THIS
\usepackage{times}  % DO NOT CHANGE THIS
\usepackage{helvet}  % DO NOT CHANGE THIS
\usepackage{courier}  % DO NOT CHANGE THIS
\usepackage[hyphens]{url}  % DO NOT CHANGE THIS
\usepackage{graphicx} % DO NOT CHANGE THIS
\usepackage{natbib}  % DO NOT CHANGE THIS AND DO NOT ADD ANY OPTIONS TO IT
\usepackage{caption} % DO NOT CHANGE THIS AND DO NOT ADD ANY OPTIONS TO IT
\usepackage{algorithm}
\usepackage{algorithmic}
\usepackage{amsmath} 
\usepackage{adjustbox}
\usepackage{wasysym}
\usepackage{multirow} 
\usepackage{xcolor}
\usepackage{amssymb}
\usepackage{booktabs}
\usepackage{makecell}

\usepackage{newfloat}
\usepackage{listings}
\DeclareCaptionStyle{ruled}{labelfont=normalfont,labelsep=colon,strut=off} % DO NOT CHANGE THIS
\floatstyle{ruled}
\newfloat{listing}{tb}{lst}{}
\floatname{listing}{Listing}
\title{Adaptive Morph-Patch Transformer for Aortic Vessel Segmentation}
\author{
    Zhenxi Zhang\textsuperscript{\rm 1,3}\thanks{Work done during the internship at SIAT.}, 
    Fuchen Zheng\textsuperscript{\rm 2}, 
    Adnan Iltaf\textsuperscript{\rm 1}, 
    Yifei Han\textsuperscript{\rm 3}, 
    Zhenyu Cheng\textsuperscript{\rm 1}, 
    Yue Du\textsuperscript{\rm 1}, 
    Bin Li\textsuperscript{\rm 1}\thanks{Corresponding authors.},
    Tianyong Liu\textsuperscript{\rm 1,4}\footnotemark[2],
    Shoujun Zhou\textsuperscript{\rm 1}\footnotemark[2]\
}
\affiliations{
    \textsuperscript{\rm 1} Institute of Scientific Instrumentation, Shenzhen Institutes of Advanced Technology, Chinese Academy of Sciences\\
    \textsuperscript{\rm 2} Department of Computer and Information Science, University of Macau\\
    \textsuperscript{\rm 3} Department of Health Technology and Informatics, The Hong Kong Polytechnic University\\
    \textsuperscript{\rm 4}  ‌School of Computer Science and Technology, Tongji University\\
    \{zx.zhang3, adnan, zy.cheng, yue.du2, b.li2, sj.zhou\}@siat.ac.cn, yc37950@um.edu.mo, 20100916d@connect.polyu.hk, tianyong@tongji.edu.cn
}

\usepackage{bibentry}
\begin{document}

\maketitle

\begin{abstract}
% Accurate segmentation of aortic vascular structures is crucial for cardiovascular disease diagnosis and treatment. 
% Traditional Transformer-based models excel in aortic vascular segmentation by capturing long-range dependencies between distant vessel features. 
% However, these models divide images into fixed-size rectangular patches, which disrupt thin and continuous vascular structures. 
% To address this limitation, we propose the adaptive Morph-Patch Transformer (MPT), a novel network designed specifically for aortic vascular segmentation. 
% MPT introduces an adaptive patch partitioning strategy that generates irregularly shaped patches aligned with the vessel morphology. 
% This strategy better preserves the semantic information within patches and alleviates vascular structure damage caused by rigid partitioning.
% Moreover, a clustering method is proposed to dynamically aggregate features from various patches with similar semantic characteristics.
% This method enables the model to capture information from different scales, improving segmentation accuracy for vascular structures of varying sizes.
% Experiments on benchmark datasets demonstrate that MPT outperforms state-of-the-art methods, particularly in segmenting thin and complex vessels. 
% Our work highlights the potential of MPT in advancing cardiovascular imaging and diagnostics.

Accurate segmentation of aortic vascular structures is critical for diagnosing and treating cardiovascular diseases. Traditional Transformer-based models have shown promise in this domain by capturing long-range dependencies between vascular features. 
However, their reliance on fixed-size rectangular patches often influences the integrity of complex vascular structures, leading to suboptimal segmentation accuracy. 
To address this challenge, we propose the adaptive Morph-Patch Transformer (MPT), a novel architecture specifically designed for aortic vascular segmentation. 
Specifically, MPT introduces an adaptive patch partitioning strategy that dynamically generates morphology-aware patches aligned with complex vascular structures.
This strategy can preserve semantic integrity of complex vascular structures within individual patches.
Moreover, a Semantic Clustering Attention (SCA) method is proposed to dynamically aggregate features from various patches with similar semantic characteristics.
This method enhances the model's capability to segment vessels of varying sizes, preserving the integrity of vascular structures.
%Additionally, we develop a semantic clustering module to aggregate features from morphologically similar patches across multiple scales, enhancing the model's ability to segment vessels of varying sizes. 
Extensive experiments on three open-source dataset(AVT, AortaSeg24 and TBAD) demonstrate that MPT achieves state-of-the-art performance, with improvements in segmenting intricate vascular structures. 
\end{abstract}
\begin{links}
    \link{Code}{https://github.com/iCherishxixixi/MPTransformer}
\end{links}


\section{Introduction}
Cardiovascular diseases (CVDs)  remain a primary cause of morbidity and mortality globally, emphasizing the necessity for precise and timely diagnosis to improve patient outcomes~\cite{townsend2022epidemiology}.
Aortic vascular segmentation, which involves delineating the structure of the aorta and its branches from medical images, plays a pivotal role in diagnosing and planning treatments for various cardiovascular conditions~\cite{hahn2021artificial}, including aneurysms, dissections, and stenosis. 
The precision of vascular segmentation directly impacts the reliability of downstream tasks~\cite{yagis2024deep}, such as computational flow modeling, surgical planning, and disease progression monitoring.
In recent years, deep learning models have revolutionized aorta segmentation, with Transformer-based architectures emerging as a dominant paradigm~\cite{lin2023deformable,li2024towards}. 
These models excel at capturing long-range dependencies and contextual information, making them well-suited for handling vascular structures, which extend across large spatial regions~\cite{dosovitskiy2020vit,zhang2024dct}.
\begin{figure}[t]
\includegraphics[width=0.5\textwidth]{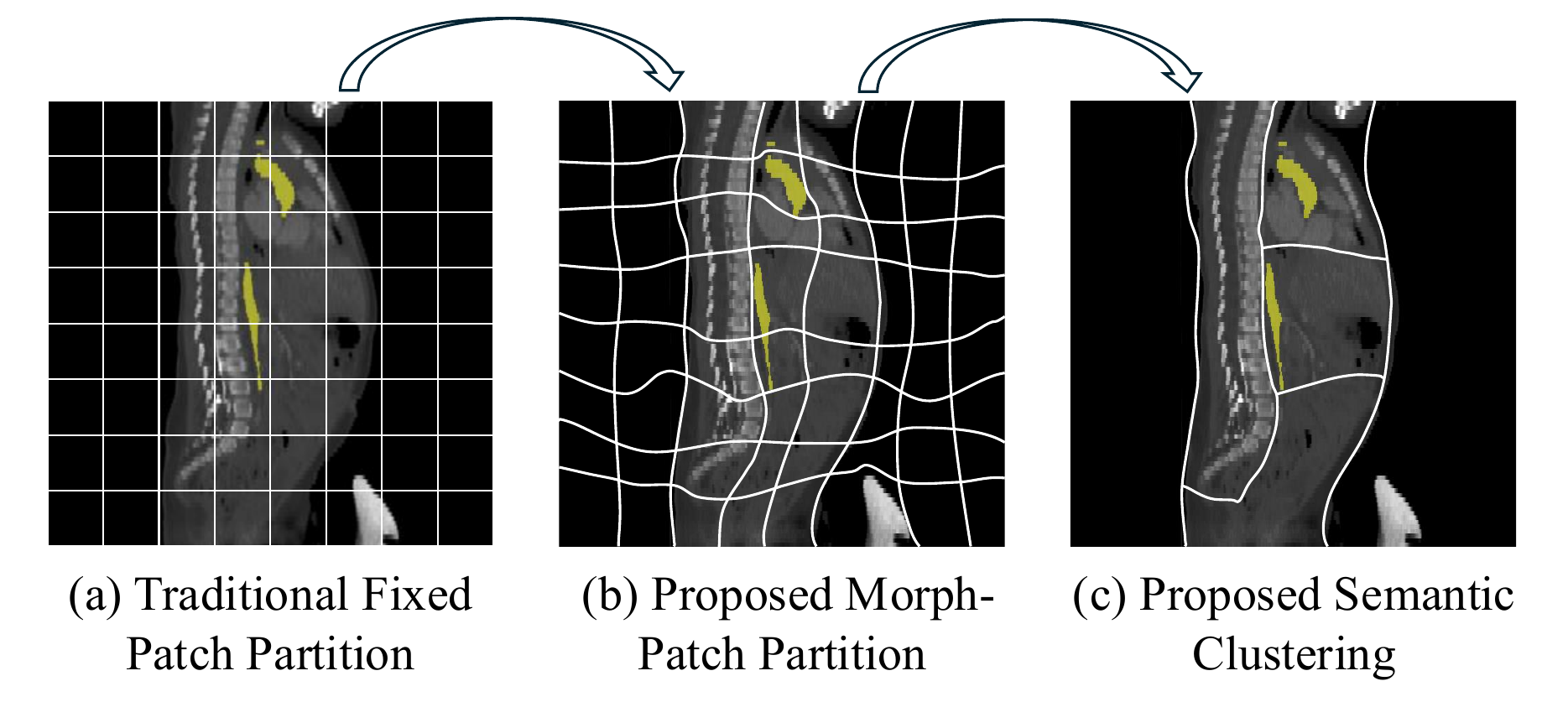}
\caption{Transformer improvements for tailoring vascular structure segmentation. (a) Traditional Transformers create fixed-size patches. (b) Our method creates morphed patches. (c) Morphed patches are grouped based on semantic similarities.} 
\label{fig1}
\end{figure}

\begin{figure*}[t]
\centering
\includegraphics[width=1\textwidth]{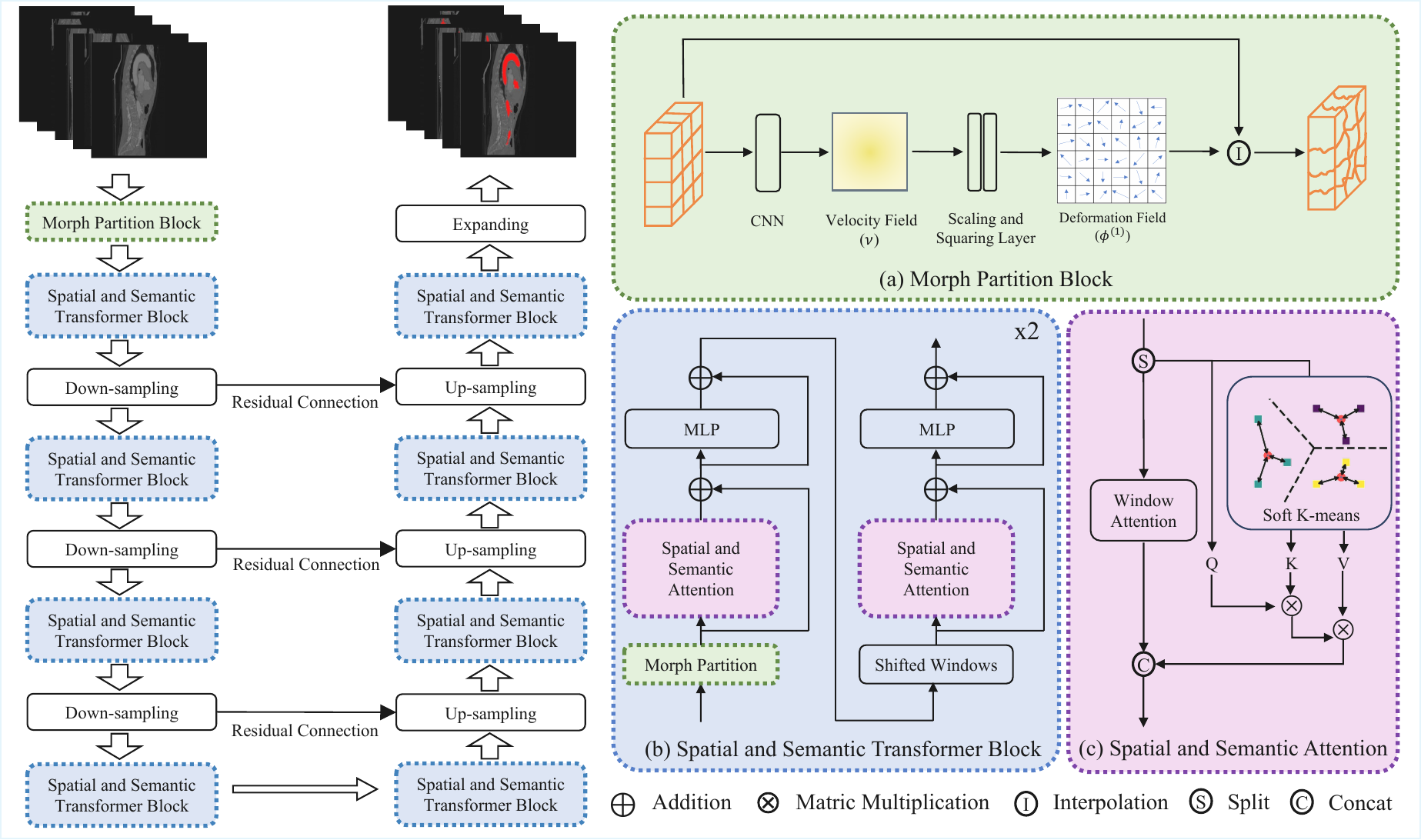}
\caption{Overall architecture of the proposed Morph-Patch Transformer for vascular structure segmentation. (a) Morph partition block to create morphology-aware patches. (b) The Transformer block with spatial and semantic attention. (c) Spatial and semantic attention is combined by window attention and semantic clustering attention.} 
\label{fig2}
\end{figure*}
However, applying traditional Transformers~\cite{hatamizadeh2022unetr,wang2021transbts,zhou2023nnformer} to vascular segmentation remains following challenges: (1) \textbf{Fixed patch partition hinders extraction of complex vascular shapes.} 
As shown in Fig.~\ref{fig1} (a), traditional Transformers divide images into fixed-size patches, which struggle to preserve the semantic integrity of intricate blood vessels.
To accommodate complex morphological structures, DCN~\cite{dai2017deformable} introduces a learnable deformation field to capture morphology-aware receptive fields in CNNs. 
Building on this, DPT~\cite{chen2021dpt} extends this method to Transformers, generating variable-sized  rectangular patches in a data-driven manner.
Although DPT is effective for natural images, it struggles to frame fragile and thin vessels within rectangular patches.
Recent Transformer-based models~\cite{zhu2024ttcnet,jian2025dau,xian2025dcfu} adopt hybrid architectures to extract features of elongated vessels.
Specifically, these Transformer-based models integrate Snake Convolution~\cite{qi2023dynamic}, which imposes prior constraints on the deformation field, to enhance feature extraction for elongated vessels.
Although these methods adapt Transformers for vessel segmentation, they still face challenges in accurately capturing fine vessel details due to rectangular patch partition.
(2) \textbf{Single-scale patches hinder the learning of semantic similarities at different scales.} 
Pyramid Transformer~\cite{zhang2020feature} enables multi-scale feature extraction with its pyramid structure, but its fixed design limits adaptability to targets of varying scales.
To address this, Swin-Transformer~\cite{liu2021swin, zhou2023nnformer} introduces hierarchical window attention, which not only captures multi-scale features more effectively but also reduces computational complexity. 
Despite these advancements, Swin-Transformer's reliance on fixed windows remains a limitation in modeling highly complex structures.
Latest studies~\cite{xia2022vision,azad2024beyond} have drawn inspiration from DCN to develop deformable attention mechanisms, enabling dynamic window adjustments. 
Nevertheless, these methods fail to incorporate semantic similarities, limiting their overall effectiveness.

%challenges:
To address these limitations, we propose the adaptive Morph-Patch Transformer (MPT), a novel model tailored for aortic vascular segmentation. 
MPT introduces an adaptive patch partitioning strategy guided by a velocity field. 
Unlike traditional deformable convolutions~\cite{dai2017deformable,qi2023dynamic}, which directly generate deformation fields, our method first constructs a velocity field and iteratively computes the final deformation shown in Fig.~\ref{fig1} (b). 
This method aligns patches more naturally with vascular structures, enabling the model to capture fine-grained vessel features without compromising topological integrity~\cite{mukherjee2015differential}. 
Furthermore, we propose a clustering attention mechanism that dynamically aggregates features from patches with similar semantic characteristics. 
This approach allows the model to effectively capture various vascular structures with semantic consistency as shown in Fig.~\ref{fig1} (c). 
By adaptively extracting vascular features at different scales, it enhances the segmentation of vessels, particularly for those with varying sizes.
Extensive experiments on benchmark datasets demonstrate that MPT achieves state-of-the-art performance, outperforming other advanced Transformer-based models. 
The proposed method significantly improves segmentation accuracy for complex vascular structures, particularly in capturing fine-grained details and preserving topological integrity. 
Our contributions are summarized as follows:
\begin{itemize}
    \item We introduce a novel \textbf{adaptive Morph-Patch Transformer (MPT)} that utilizes an adaptive patch partitioning strategy on a velocity field. This method effectively preserves vascular topology and enhances the alignment of patches with intricate blood vessel structures.
    %%%1. Semantic Cluster Attention (SCA)
    % 该方法的核心是通过 Soft K-means 对特征进行语义聚类，并利用聚类中心进行注意力计算，"Semantic Cluster" 强调了聚类的语义建模能力，"Attention" 则体现了注意力机制的融合。
%     2. Adaptive Semantic Clustering Transformer (ASCT)
% 命名原因：强调了该方法的自适应性（Adaptive）、语义聚类（Semantic Clustering）和 Transformer 框架的结合。聚类中心是动态更新的，体现了自适应特性。
% 3. Morphological Semantic Attention (MSA)
% 命名原因：结合了方法的目标（处理复杂形态结构，如血管）和语义注意力的核心机制，"Morphological" 强调其在医学图像等复杂形态结构中的适用性。
% 4. Soft Cluster Attention Mechanism (SCAM)
% 命名原因：直接体现了 Soft K-means 聚类和注意力机制的结合，简洁明了。
% 5. Dynamic Cluster-Aware Transformer (DCAT)
% 命名原因：强调动态聚类（Dynamic Cluster）和 Transformer 框架的结合，体现了方法的动态性和全局建模能力。
    \item We propose a \textbf{Semantic Clustering Attention (SCA)} method that dynamically aggregates features from patches with similar semantic characteristics, enhancing the segmentation of vessels across varying sizes.
    \item We conduct extensive experiments on three open-source  dataset(AVT, AortaSeg24 and TBAD), demonstrating that MPT achieves \textbf{State-of-the-Art} performance, particularly in segmenting intricate vascular structures.
\end{itemize}

\section{Proposed Method}
\subsection{Overall Morph-Patch Transformer Architecture}
In this work, we propose the Morph-Patch Transformer (MPT), a novel architecture designed to address two key challenges in vascular structure segmentation. 
First, MPT tackles the complex and irregular morphological structures of vessels through morphology-aware patch partitioning, which adapts to the inherent geometry of vascular networks.
Second, MPT addresses the need to capture multi-scale semantic features by introducing fusion attention, a mechanism that models both spatial and semantic contextual relationships across different scales. 
Built on a 3D UNet-like framework~\cite{cciccek20163d}, MPT combines these innovations to achieve precise and robust segmentation. 
The key components are illustrated in Fig.~\ref{fig2} and detailed below.
\begin{itemize}
    \item \textbf{Morph Partition Block:} This block first utilizes a CNN to predict the velocity field shown in Fig.~\ref{fig2} (a).
    The predicted velocity field is then transformed into a diffeomorphic deformation field through the scaling and squaring method~\cite{arsigny2006log}, where each point in the field represents coordinate offsets. 
    This diffeomorphic nature not only guarantees smooth and invertible transformations but also naturally preserves the continuity of vascular structures throughout the deformation process. 
    Finally, the coordinate offsets are used to generate deformed features through bilinear interpolation of the original input. 
    From these transformed features, the method extracts patches that effectively capture complex vessel structures while preserving their topological relationships.
    \item \textbf{Spatial and Semantic Transformer Block:} This block is designed to effectively integrate spatial and semantic contextual relations for vascular feature learning. 
    As shown in Fig.~\ref{fig2} (b), the module builds upon SwinTransformer's~\cite{liu2021swin} window attention and shift window strategies to capture spatial relationships. 
    To address the complexity of vascular structures, we introduce the Morph Partition Block (Fig.~\ref{fig2} (a)), which dynamically adapts window shapes to fit varying morphological patterns. 
    Furthermore, to enhance semantic understanding, Semantic Clustering Attention (SCA) (Fig.~\ref{fig2} (c)) is proposed, where a soft-Kmeans algorithm extracts key semantics and computes their relationships with patch features. 
    These components are integrated into a 3D UNet-based framework~\cite{cciccek20163d}, enabling multi-scale fusion of spatial and semantic relations. 
\end{itemize}

\subsection{Morph-Patch Feature} 
The morph-patch feature is derived from the morph partition block, as detailed in \textbf{Algorithm~\ref{alg:morph_patch}}. 
This feature is designed to extract complex vascular structures and is formulated as:
\begin{algorithm}[tb]
\caption{Morph-Patch Feature Extraction with Diffeomorphic Deformation Field}
\label{alg:morph_patch}
\textbf{Input:} Feature map $x$, patch center $p_0$, neighborhood region $\mathcal{R}$, velocity field $\upsilon$, number of steps $n$\\
\textbf{Parameter:}  Spatial weighting function $w$\\
\textbf{Output:} Morph-patch feature $y(p_0)$\\
\begin{algorithmic}[1]
\STATE Let $\Delta t \gets \frac{1}{2^n}$
\STATE Initialize deformation field $\phi^{(0)} \gets Id$
\STATE Compute initial deformation: $\phi^{(\Delta t)} \gets (Id + \upsilon(\Delta t)) \circ \phi^{(0)}$
\FOR{$i = n$ to $1$}
    \STATE $\phi^{\left(\frac{1}{2^{i-1}}\right)} \gets \phi^{\left(\frac{1}{2^i}\right)} \circ \phi^{\left(\frac{1}{2^i}\right)}$
\ENDFOR
\STATE Set final deformation field $\phi \gets \phi^{(1)}$
\STATE Initialize $y(p_0) \gets 0$
\FOR{each $p_n \in \mathcal{R}$}
    \STATE $y(p_0) \gets y(p_0) + w(p_n) \cdot x(p_0 + p_n + \phi(p_0 + p_n))$
\ENDFOR
\STATE \textbf{return} $y(p_0)$
\end{algorithmic}
\end{algorithm}
\begin{equation}
y(p_0) = \sum_{p_n \in \mathcal{R}} w(p_n) \cdot x(p_0 + p_n + \phi(p_0 + p_n)),
\end{equation}
where $w$ is the feature weight, $p_0$ represents the center of a patch, $p_n$ denotes its neighboring regions within the patch region $\mathcal{R}$, and $\phi(p_0 + p_n)$ represents the deformation field, which provides coordinate offsets to adaptively adjust the sampling locations.
To ensure that the transformation preserves the topological properties of the features before and after deformation, we employ a stationary velocity field $\upsilon $ to generate a diffeomorphic mapping. 
The velocity field $\upsilon $ is integrated over time $t=[0,1]$ to obtain the final deformation field $\varphi ^{(1)} $, as described by the Ordinary Differential Equation (ODE):
\begin{equation}
\frac{\partial \phi^{(t)}  }{\partial t} =\upsilon (\phi^{(t)}),~~~with~\phi^{(0)} = Id,
\end{equation}
%建议明确 $\upsilon$ 的性质（如是否连续、是否光滑），以确保 $\phi^{(t)}$ 是光滑的微分同胚（diffeomorphism）。连续时间的 ODE建议补充如何选择时间步长 $\Delta t$。
where $Id$ represents the identity transformation. 
This formulation ensures that $\phi^{(t)}$ is a diffeomorphism, meaning it is smooth, invertible, and preserves the topological structure of the input features.

The deformation field $\phi^{(1)}$ is computed numerically using the scaling and squaring method~\cite{dalca2019unsupervised}. This approach leverages the properties of one-parameter subgroups of diffeomorphisms, where the velocity field $\upsilon$ lies in the Lie algebra and is exponentiated to generate the deformation field $\phi^{(1)}$ in the Lie group. 
To ensure numerical stability and accuracy, the computation begins by calculating the deformation field such that $t$ is sufficiently small ($t = \frac{1}{2^{n} } \approx 0$). With the scaled velocity field, the initial deformation field is represented as:
\begin{equation}
\phi^{(\frac{1}{2^{n} } )}= (Id+\upsilon (\frac{1}{2^{n} })) \circ \phi^{(0)}, 
\end{equation}
where $\circ$ represents the composition operation. The deformation field is then iteratively refined using the recurrence relation:
\begin{equation}
\phi^{(\frac{1}{2^{n-1} } )} = \phi^{(\frac{1}{2^{n} } )} \circ \phi^{(\frac{1}{2^{n} } )}.
\end{equation}

By constructing $\phi^{(1) }$ in this manner, the transformation inherently preserves the topological integrity of the input features. 
This is particularly crucial in applications such as medical image analysis, where maintaining the structural continuity of complex vascular features is essential.
\subsection{Semantic Clustering Attention with Soft K-means } 
In this section, we propose a Soft K-means module for semantic clustering, which extracts core semantic features to model semantic relationships in the data. 
The module is formulated as follows:
\begin{equation}
f_{newcore}^{s} = {\textstyle \sum_{i=1}^{m}} g_{s} (f^{i} ) d(f^{i},f_{core}^{s}),
\end{equation}
% 公式中 $d(f(i), f_{core}(s))$ 未明确。如果是欧氏距离，建议标注为 $d(f(i), f_{core}(s)) = |f(i) - f_{core}(s)|^2$。

\noindent where $F = \left \{ f^{1},f^{2}, ..., f^{m} \right \}$ represents all patch features, 
$F_{core} = \{ f_{core}^{1}$, $f_{core}^{2}$, $...,$ $f_{core}^{n} \}$ denotes the original core semantic features, $F_{newcore}= \ \{f_{newcore}^{1}$, $f_{newcore}^{2}, ..., f_{newcore}^{n} \}$  represents the new core semantic features, $g_{s}(\cdot)$ reflects the importance of all features $F$ to the original core semantic feature $f_{core}^{s}$, and $d(\cdot,\cdot ) $ measures the similarity between two features.
In conventional K-means algorithm, $g_{s}(\cdot)$ would be a non-differentiable discrete function. 
Specifically, $g_{s}(\cdot)$ would be 1 if $f^{i}$ belongs to the cluster corresponding to $f_{core}^{s}$, and 0 otherwise. 
To ensure differentiability, we design a smoothed version of $g_{s}(\cdot)$:
\begin{equation}
 g_{s} (f^{i})= \frac{e^{-\beta {\left \| f^{i}-f_{core}^{s} \right \|}^{2}}  }{ {\textstyle \sum_{k=1}^{n}} e^{-\beta {\left \| f^{i}-f_{core}^{k} \right \|}^{2}}} ,
\end{equation}
By defining $\lambda^{s} = 2\beta f_{core}^{s}$,  $\mu^{s} = -\beta \left \| f_{core}^{s} \right \|^{2}$, and $d(\cdot,\cdot ) $ as vector subtraction, $f_{core}^{s}$ can be simplified to:
% 缺少补充 $\lambda(s)$ 和 $\mu(s)$ 的物理意义，例如它们如何影响聚类中心的更新。
\begin{equation}
f_{newcore}^{s}=  {\textstyle \sum_{i=1}^{m}} \frac{e^{\lambda^{s} f^{i}+\mu^{s} }  }{ {\textstyle \sum_{k=1}^{n}} e^{\lambda^{k} f^{i}+\mu^{k}}}(f^{i}-f_{core}^{s}).
\end{equation}
In practice, $\lambda$, $\mu$ and $F_{core}$ are learned by the neural network.
The updated semantic centers $f_{newcore}$ are then applied to compute Semantic Clustering Attention (SCA), formulated as:
\begin{align}
\text{SCA}(F) &= \text{softMax} \left(
    \frac{(F W_Q)(F_{\text{newcore}} W_K)^T}{\sqrt{d}}
\right) \nonumber \\
&\quad \cdot (F_{\text{newcore}} W_V) \label{eq:sca}
\end{align}
where $W_{Q}$, $W_{K}$ and $W_{V}$ are learnable weight matrices. 
This attention mechanism enables the model to effectively integrate semantic relationships, enhancing the model's ability to capture complex patterns in the data.
\begin{figure}[h!]
\centering
\includegraphics[width=0.48\textwidth]{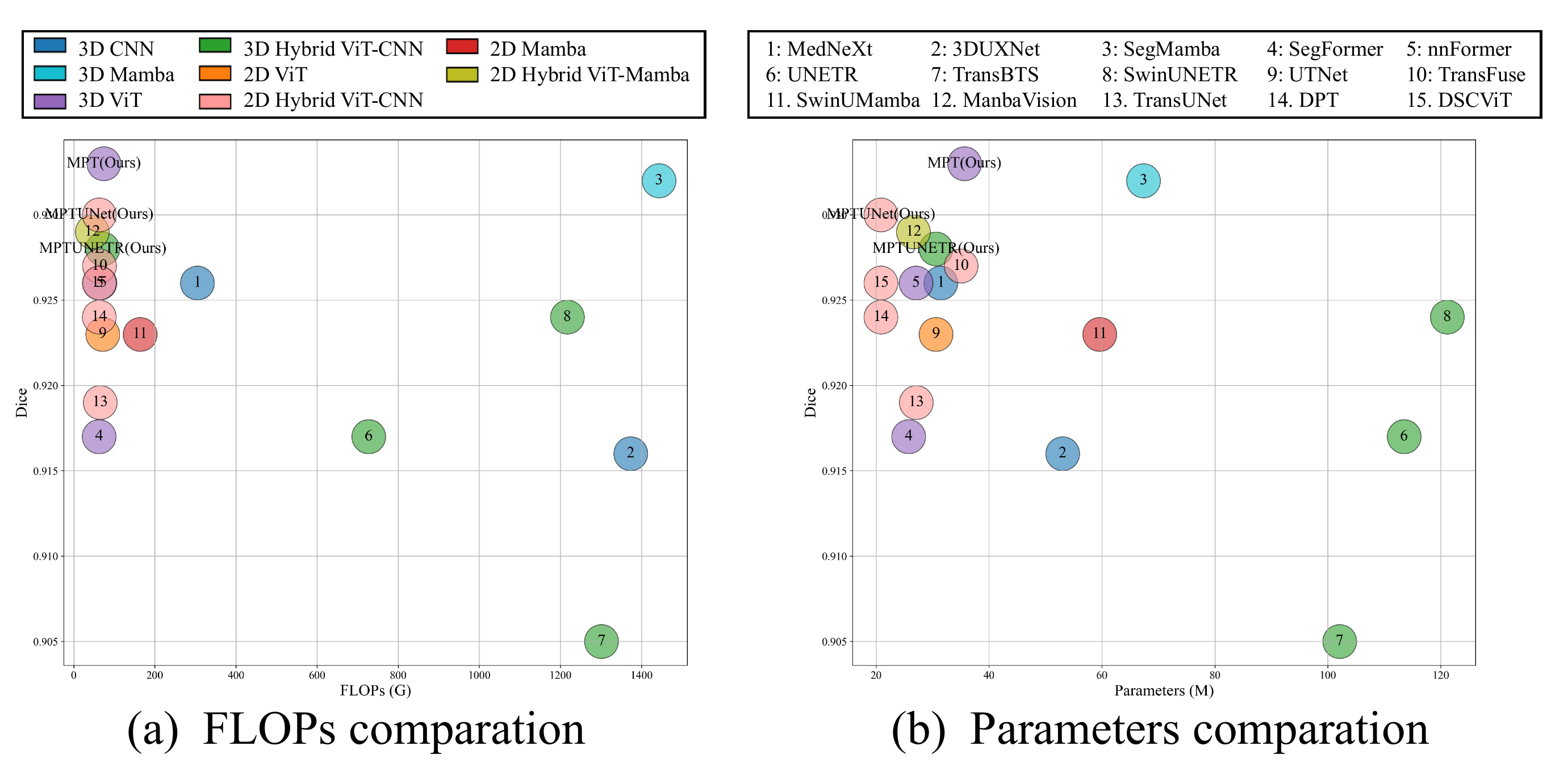}
\caption{Accuracy–Efficiency Comparison on TBAD: (a) FLOPs, (b) Parameters; Colors Indicate Backbone Types.} 
\label{fig3}
\end{figure}
\begin{table*}
\centering
\begin{tabular}{c|c|c|c|c|c|c} 
\hline
\multirow{2}{*}{Model}  & \multicolumn{3}{c|}{AVT~}                                             & \multicolumn{3}{c}{TBAD~}                                              \\ 
\cline{2-7}
                        & Dice                  & mIoU                  & clDice                & Dice                  & mIoU                  & clDice                 \\ 
\hline
MedNeXt~                & 0.809(0.198)          & 0.718(0.233)          & 0.724(0.185)          & 0.926(0.172)          & 0.871(0.190)          & 0.880(0.072)           \\ 
3DUXNet~                & 0.843(0.095)          & 0.740(0.136)          & 0.745(0.105)          & 0.916(0.187)          & 0.859(0.207)          & 0.892(0.042)           \\ 
SegMamba ~              & 0.829(0.134)          & 0.730(0.179)          & 0.711(0.177)          & 0.932(0.164)          & 0.881(0.178)          & 0.918(0.039)           \\ 
SegFormer ~             & 0.837(0.124)          & 0.737(0.159)          & 0.770(0.136)          & 0.912(0.168)          & 0.854(0.181)          & 0.893(0.062)           \\ 
nnFormer ~              & 0.835(0.158)          & 0.743(0.201)          & 0.732(0.169)          & 0.926(0.158)          & 0.871(0.173)          & 0.895(0.039)           \\ 
\textbf{MPT(Ours)}      & 0.856(0.106)          & 0.762(0.150)          & 0.757(0.152)          & \textbf{0.933(0.150)} & \textbf{0.881(0.161)} & 0.915(0.032)           \\ 
UTNETR~                 & 0.791(0.145)          & 0.677(0.181)          & 0.674(0.150)          & 0.917(0.157)          & 0.864(0.170)          & 0.901(0.044)           \\ 
TransBTS ~              & 0.822(0.149)          & 0.722(0.190)          & 0.770(0.163)          & 0.905(0.172)          & 0.841(0.191)          & 0.899(0.050)           \\ 
SwinUNETR~              & 0.811(0.224)          & 0.729(0.248)          & 0.740(0.222)          & 0.924(0.187)          & 0.870(0.204)          & 0.906(0.060)           \\ 
\textbf{MPTUNETR(Ours)} & 0.829(0.162)          & 0.736(0.203)          & 0.764(0.180)          & 0.928(0.159)          & 0.874(0.174)          & 0.911(0.041)           \\ 
\hline
UTNet ~                 & 0.861(0.072)          & 0.762(0.106)          & 0.780(0.087)          & 0.923(0.150)          & 0.865(0.161)          & 0.911(0.025)           \\ 
TransFuse~              & 0.880(0.079)          & 0.794(0.115)          & 0.796(0.107)          & 0.927(0.139)          & 0.872(0.145)          & 0.895(0.063)           \\ 
SwinUMamba~             & 0.861(0.067)          & 0.761(0.100)          & 0.771(0.093)          & 0.923(0.150)          & 0.867(0.158)          & 0.902(0.056)           \\ 
MambaVision~            & 0.882(0.064)          & 0.795(0.097)          & 0.795(0.099)          & 0.929(0.148)          & 0.874(0.151)          & 0.914(0.037)           \\ 
TransUNet~              & 0.874(0.064)          & 0.782(0.093)          & 0.801(0.091)          & 0.919(0.146)          & 0.860(0.159)          & 0.894(0.051)           \\ 
DPT ~                   & 0.886(0.055)          & 0.800(0.086)          & 0.825(0.089)          & 0.924(0.137)          & 0.868(0.144)          & 0.917(0.026)           \\ 
DSCViT~                 & 0.877(0.060)          & 0.786(0.090)          & 0.782(0.085)          & 0.926(0.151)          & 0.870(0.159)          & 0.893(0.062)           \\ 
\textbf{MPTUNet(Ours)}  & \textbf{0.896(0.046)} & \textbf{0.815(0.073)} & \textbf{0.839(0.078)} & 0.930(0.147)          & 0.877(0.156)          & \textbf{0.920(0.031)}  \\
\hline
\end{tabular}
\caption{Experimental results of aortic segmentation on AVT and TBAD dataset. Values are reported as ``mean (standard deviation)". }
\label{tab1}
\end{table*}
\section{Experiments}
\begin{table}[t]
\centering
\setlength{\tabcolsep}{3pt} 
\begin{tabular}{lccc}
\toprule
Model & Dice & mIoU & clDice \\
\midrule
MedNeXt & 0.758 (0.077) & 0.631 (0.080) & 0.963 (0.018) \\
3DUXNet & 0.784 (0.077) & 0.666 (0.083) & 0.964 (0.011) \\
SegMamba & 0.747 (0.076) & 0.620 (0.080) & 0.924 (0.022) \\
SegFormer & 0.753 (0.104) & 0.630 (0.105) & 0.951 (0.023) \\
nnFormer & 0.779 (0.056) & 0.666 (0.056) & 0.923 (0.025) \\
\textbf{MPT} & \underline{0.804 (0.021)} & \underline{0.690 (0.024)} & 0.926 (0.032) \\
UTNETR & 0.753 (0.096) & 0.630 (0.099) & 0.916 (0.024) \\
TransBTS & 0.721 (0.078) & 0.594 (0.075) & 0.928 (0.040) \\
SwinUNETR & 0.781 (0.091) & 0.664 (0.097) & 0.937 (0.027) \\
\textbf{MPTUNETR} & \textbf{0.809 (0.045)} & \textbf{0.695 (0.054)} & 0.955 (0.015) \\
\midrule
UTNet & 0.715 (0.056) & 0.593 (0.059) & 0.905 (0.035) \\
TransFuse & 0.739 (0.054) & 0.630 (0.057) & 0.890 (0.047) \\
SwinUMamba & 0.766 (0.080) & 0.648 (0.083) & 0.961 (0.017) \\
MambaVision & 0.795 (0.071) & 0.682 (0.076) & 0.960 (0.021) \\
TransUNet & 0.751 (0.103) & 0.633 (0.097) & 0.922 (0.045) \\
DPT & 0.778 (0.075) & 0.662 (0.079) & 0.959 (0.017) \\
DSCViT & 0.788 (0.077) & 0.673 (0.077) & 0.965 (0.019) \\
\textbf{MPTUNet} & 0.796 (0.075) & 0.686 (0.077) & \textbf{0.966 (0.011)} \\
\bottomrule
\end{tabular}
\caption{Experimental results of aortic segmentation on AortaSeg24 dataset.  Values are reported as ``mean (standard deviation)".}
\label{tab2}
\end{table}

\subsection{Dataset and Implementation} 
In this section, we evaluate the efficacy of MPT on three widely used open-source aorta datasets: AVT~\cite{radl2022avt}, TBAD~\cite{yao2021imagetbad}, and AortaSeg24~\cite{imran2025multi}.
The AVT dataset contains 56 cases and focuses on a single-class segmentation task, requiring the extraction of the aorta from CTA scans collected from three hospitals. 
TBAD includes 100 high-resolution 3D CTA images of Type-B Aortic Dissection (TBAD), with detailed annotations of the true lumen (TL), false lumen (FL), and false lumen thrombus (FLT).
These annotations test the model’s ability to identify diseased aortic structures.
AortaSeg24 is one of the most detailed aorta segmentation datasets, encompassing 100 CTA scans annotated with 23 clinically meaningful aortic regions. 
Segmenting numerous and tiny vascular anatomical structures poses significant challenges.
As such, AortaSeg24 serves as a robust benchmark for evaluating the fine-grained anatomical segmentation capabilities of aortic segmentation models.

In our experiments, all models are implemented in PyTorch and trained on NVIDIA GeForce RTX 3090 GPUs with Ubuntu 20.04. 
To ensure fair and accurate comparison, we adopt the officially released codes of the referenced methods and conduct all experiments within the nnU-Net framework~\cite{isensee2021nnu}. 
Specifically, we follow the preprocessing steps of nnU-Net, including normalization, resampling, and cropping. 
Each dataset is split into training, validation, and test sets in an 8:1:1 ratio, and image sizes are standardized to 128×128×128 for 3D and 512×512 for 2D to ensure consistent input dimensions. 
To verify the effectiveness of our proposed models, we provide three versions: MPT, a pure 3D ViT-based architecture; MPT-UNETR, a hybrid 3D ViT-CNN framework; and MPT-UNet, a lightweight 2D model. 
Additionally, the number of clusters is set to 32 in these networks.
All three models are optimized using the Adam optimizer with a learning rate of $5 \times 10^{-5}$. Training follows the nnU-Net strategy~\cite{isensee2021nnu} and is terminated after 1000 epochs. 
The Dice coefficient is employed as the loss function to guide segmentation performance.

% \vspace{-0.5cm}
%\subsection{Result} 
%The results are shown in \cref{tab1,tab2}
\begin{figure*}[h!]
\centering
\includegraphics[width=\textwidth]{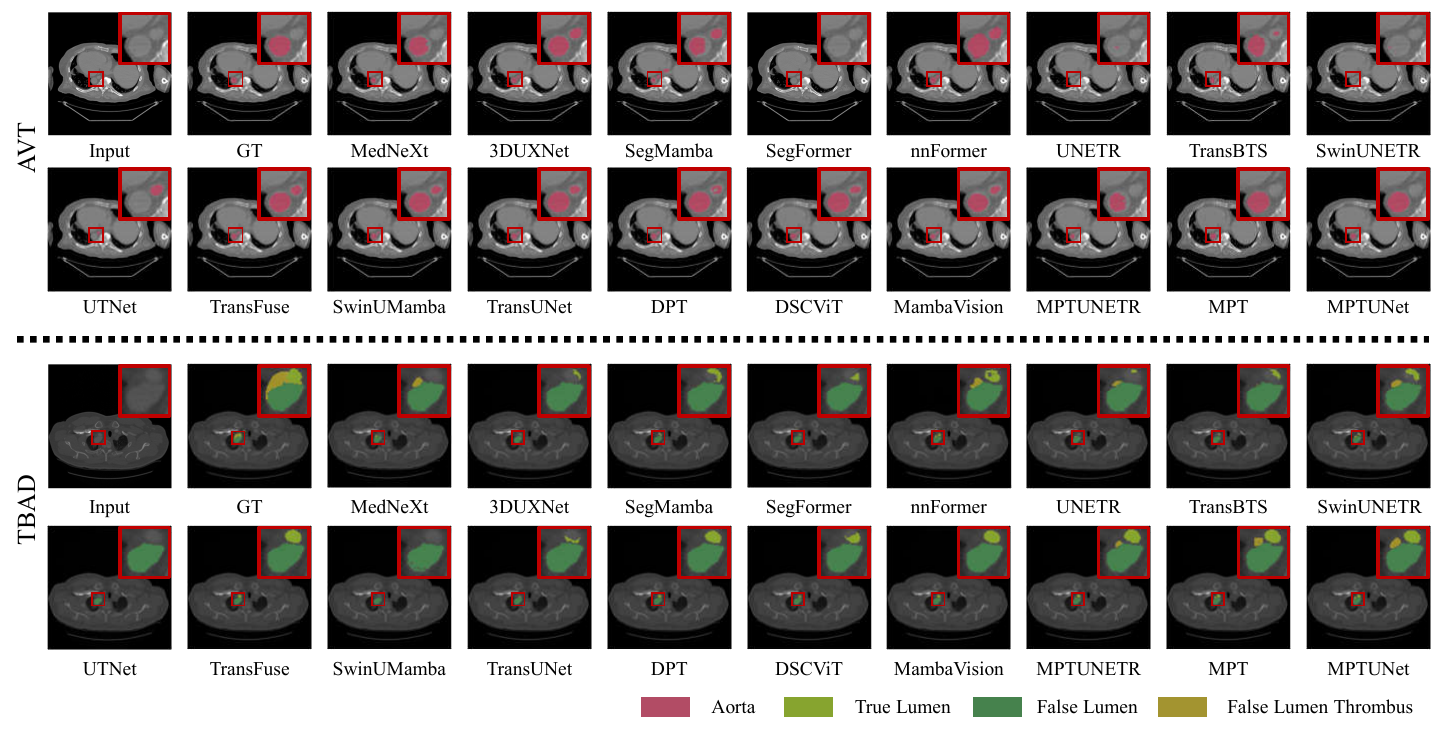}
\caption{Visual results on the AVT and TBAD datasets. The first two rows illustrate the aortic segmentation performance of different models on the AVT dataset, while the bottom two rows present their ability to identify aortic dissection on the TBAD dataset. Red-boxed areas have been magnified to better visualize specific anatomical structures.} 
\label{fig4}
\end{figure*}
\begin{figure}[h!]
\centering
\includegraphics[width=0.475\textwidth]{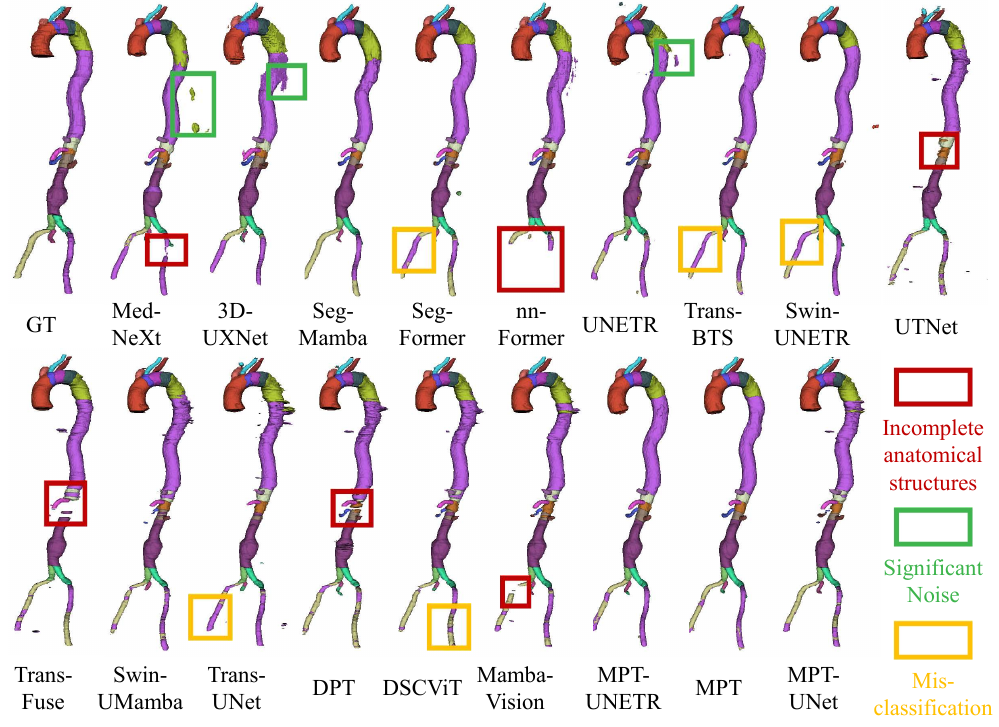}
\caption{Comparison of different models in processing complex vascular structures on the AortaSeg24 dataset. Red boxes denote incomplete anatomical structures, green boxes highlight significant noise, and yellow boxes indicate misclassifications.} 
\label{fig5}
\end{figure}
\subsection{Baselines}
%backbones
To demonstrate the superiority of our proposed models in segmenting complex aortic morphological structures, we compare MPTs with a diverse range of recent and competitive segmentation approaches. These methods are categorized into eight groups based on their backbone architectures, as illustrated in Fig.~\ref{fig3}.
Among them, 3D Vision Transformer (ViT)-based models—including SegFormer~\cite{perera2024segformer3d}, nnFormer~\cite{zhou2023nnformer}, and the proposed MPT—represent a prominent and widely studied category. 
These models tokenize volumetric inputs and employ self-attention mechanisms to capture long-range dependencies and global context effectively. 
However, pure ViT architectures typically demand significant computational resources and large-scale data to learn inductive biases effectively~\cite{lavie2024towards}. 
To mitigate these limitations, several methods adopt hybrid ViT-CNN designs to improve training efficiency and model generalizability in the 3D setting. 
Notable examples include SwinUNETR~\cite{he2023swinunetr} and our MPTUNETR, which integrate convolutional priors to enhance spatial learning while maintaining ViT-based global modeling.
On the 2D side, TransUNet~\cite{chen2021transunet} serves as a representative hybrid ViT-CNN framework, wherein convolutional blocks extract local features and Transformer encoders capture global context. 
Recent developments such as DPT~\cite{chen2021dpt} and DSCViT~\cite{qi2023dynamic} further improve segmentation performance by incorporating deformable patch embedding and dynamic snake convolution, respectively. 
These architectural advances enhance the model’s ability to delineate intricate, tubular anatomical structures. 
To verify the effectiveness of our 2D MPTUNet, we include these 2D hybrid models in our comparative evaluations.
Besides, we also incorporate the Mamba family of models in our benchmarking. 
Approaches such as SegMamba~\cite{xing2024segmamba}, SwinUMamba~\cite{liu2024swin}, and latest MambaVision~\cite{hatamizadeh2025mambavision} leverage state space models (SSMs) to efficiently model long-range dependencies while maintaining lower computational complexity compared to self-attention-based alternatives. 
This makes them particularly competitive under resource-constrained scenarios.

In Fig.~\ref{fig3}, we further present a comparison of the parameter counts and computational costs across all baseline models. 
It is evident that our proposed MPT-based models achieve superior segmentation performance while maintaining significantly lower computational complexity and fewer parameters compared to many recent state-of-the-art methods.
In addition to the Dice scores reported in Fig.~\ref{fig3}, we utilize clDice\cite{shit2021cldice}, a topology-aware metric designed to evaluate vascular structural connectivity, on several aortic datasets.
\subsection{Experiment}
%summary
The experimental results in Table~\ref{tab1} and Table~\ref{tab2} demonstrate that the proposed MPT-based methods consistently achieve superior aortic segmentation performance compared to existing 2D and 3D segmentation models. 
Specifically, among 3D approaches, the ViT-based MPT model achieves the highest scores on the AVT dataset (Dice 0.856, mIoU 0.762) shown in Table~\ref{tab1}. 
MPT also performs competitively on AortaSeg24 and TBAD, surpassing established 3D CNNs, ViT, and Mamba-based models. 
Additionally, the hybrid MPTUNETR further enhances the performance of other 3D hybrid models,including latest SwinUNETR and TransBTS. 
Notably, on the AortaSeg24 dataset, which involves complex vascular structures, MPTUNETR achieves the best Dice (0.809) and mIoU (0.695) among all methods. 
%2D
For 2D architectures, MPTUNet achieves state-of-the-art results, with Dice 0.896, mIoU 0.815, and clDice 0.839 on the AVT dataset. 
Additionally, MPTUNet outperforms other 2D segmentation models, achieving Dice scores of 0.796 and 0.930 on AortaSeg24 and TBAD, respectively. 
These results highlight the excellent generalization ability of MPTUNet. 
When compared to previous hybrid models using deformable convolutions and TransUNet (including DPT and DSCViT), MPTUNet proves more effective for handling complex tubular structures on three datasets. 
Remarkably, MPTUNet achieves the highest clDice scores across all three datasets, with values of 0.839, 0.966, and 0.920 on AVT, AortaSeg24, and TBAD, respectively. 
This confirms that the proposed morph patch strategy effectively preserves the topology and continuity of the aortic structure.
These results highlight the high performance and strong generalization capability of MPT-based models in aortic segmentation tasks.
\begin{figure}[!t]
\centering
\includegraphics[width=0.45\textwidth]{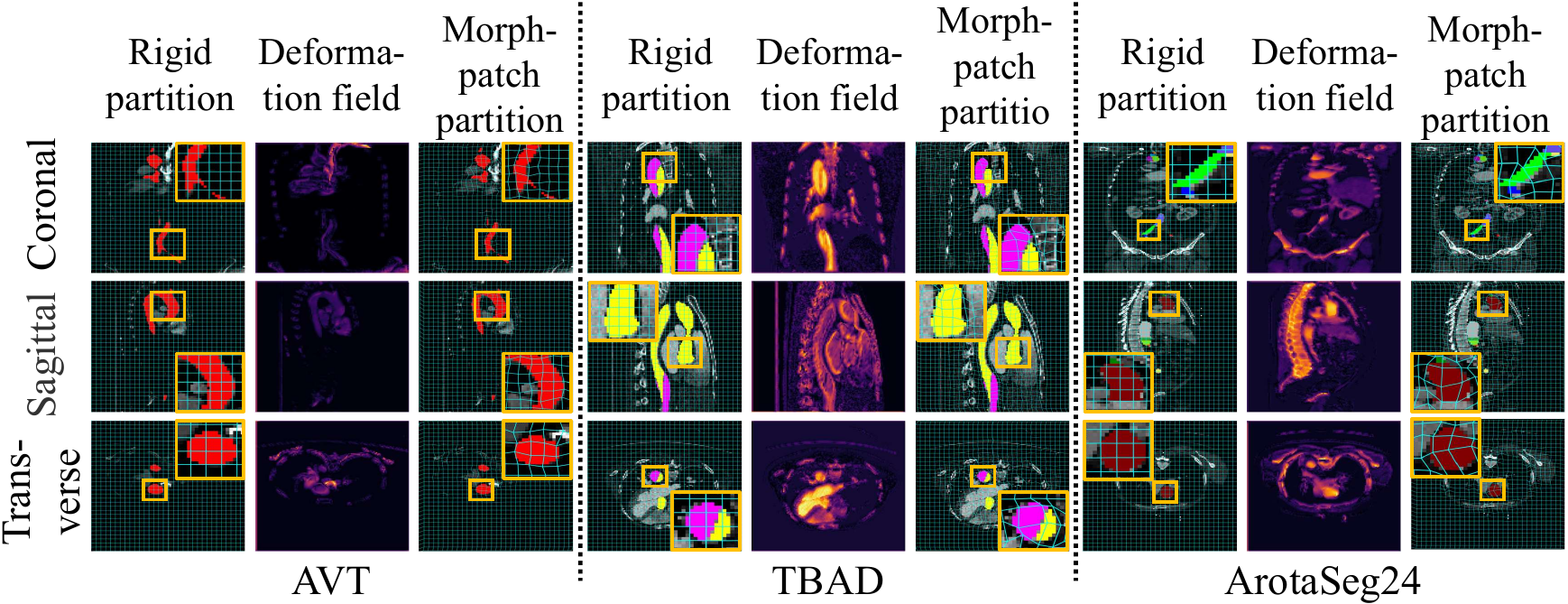}
\caption{Morph-Patch Strategy preserves vascular morphology and enables cross-view anatomical alignment.} 
\label{fig6}
\end{figure}
Fig.~\ref{fig4} and Fig.~\ref{fig5} present the visual results of MPT-based methods across three datasets. 
To be specific, the first two rows of Fig.~\ref{fig4} show the visualization results of various models on the AVT and TBAD datasets. 
The MPT-based model successfully segments the aortic structure in the AVT dataset while effectively avoiding interference from surrounding small tissue. 
The last two rows of Fig.~\ref{fig4} display the performance of different models on the TBAD dataset. 
The results indicate that only nnFormer, UNETR, SwinUNETR, and the proposed MPT-based methods are capable of displaying three anatomical structures. 
Furthermore, the MPT-based methods offer more complete segmentation results, particularly for the True Lumen, compared to the other models.
Fig.~\ref{fig5} shows the performance of different models on the AortaSeg24 dataset. 
It can be observed that the MPT-based methods handle complex vascular shapes more effectively, significantly reducing the occurrence of incomplete anatomical structures highlighted in the red boxes. 
Additionally, the MPT-based methods use SCA to aggregate similar semantic features. 
By obtaining representative features, these methods enhance the model's ability to accurately identify relevant structures, thereby reducing classification errors and minimizing segmented noise.
\subsection{Ablation Study} 

\begin{table}[h!]
\centering
\setlength{\tabcolsep}{5.5pt} 
\begin{tabular}{c|c|c|c|c|c|c}
\Xhline{1px}
\textbf{ViT} & \textbf{Mamba} & \textbf{MP} & \textbf{SCA} & \textbf{Dice} & \textbf{mIoU} &\textbf{clDice}\\
\hline
$\checkmark$ &$-$ &$-$ &$-$ & 0.834  & 0.743 & 0.732 \\
$\checkmark$ &$-$ &$\checkmark$ &$-$ & 0.839 & 0.748 & 0.744 \\
$\checkmark$ &$-$ &$\checkmark$ &$\checkmark$ & 0.856 & 0.762 & 0.757 \\
$-$ &$\checkmark$ &$-$ &$-$ & 0.829  & 0.730 & 0.711 \\
$-$ &$\checkmark$ &$\checkmark$ &$-$ & 0.831  & 0.730 & 0.725 \\
\Xhline{1px}
\end{tabular}
\captionof{table}{Ablation experiments of MPT on the AVT dataset. The MP and SCA are respectively brief expression for Morph-patch strategy and Semantic clustering attention. }
\label{tab3}
\end{table}
Table~\ref{tab2} presents the ablation experiments of MPT on the AVT dataset, where MP (Morph-Patch strategy) and SCA (Semantic Clustering Attention) are key components. 
Without MP and SCA, the Transformer-based model achieves a Dice score of 0.834, an mIoU of 0.743, and a clDice of 0.732.
Introducing MP significantly improves segmentation performance, with both Dice and mIoU scores showing noticeable gains. 
This indicates that the MP strategy adapts to the complex morphology of vessels and mitigates the structural damage caused by rigid patch partitioning. 
Moreover, the increase in clDice to 0.744 with MP further demonstrates MP's effectiveness in preserving the topological structure during vessel segmentation.
When SCA is subsequently added, the model's ability to aggregate similar semantics is enhanced, further boosting performance. 
At this point, the model achieves a Dice score of 0.856, an mIoU of 0.762, and a clDice of 0.757.

To demonstrate the generalization ability of our proposed MP strategy across different frameworks, we integrate MP with the latest Mamba architecture~\cite{liu2024swin}, which also requires patch partitioning. 
Experimental results show a significant improvement in clDice, from 0.711 to 0.725, upon introducing MP. 
This indicates that the MP strategy effectively preserves the aortic topological structures in the Mamba framework.

\subsection{Model  Analysis}
Our Morph-Patch strategy enables patches to effectively perceive complex vascular morphologies, as visualized in Fig.~\ref{fig6}. 
In the coronal view of the AortaSeg24 case, the deformation field effectively guides the patch boundaries to closely follow the contours of slender vessels. 
This results in better preservation of fine morphological details, particularly at the vessel edges.

Additionally, the adaptively generated deformation field enables the model to perceive multiple anatomical structures, including both vertebrae and vessels. 
An inspection of nine representative deformation fields from three different datasets reveals that the deformations primarily concentrate around anatomically salient regions—especially the spine and major vascular structures. 
This behavior reflects a meaningful spatial bias: in the thoracoabdominal region, the aorta and its major branches typically run adjacent to the vertebral column. 
As a result, deformation fields that consistently concentrate in these areas suggest that the model is leveraging the stable spatial relationship between the spine and major vessels as an anatomical prior. 
This implicit guidance allows the model to better localize and delineate vascular structures by referencing nearby, morphologically salient landmarks. 
Such structure-aware deformation improves not only segmentation accuracy but also interpretability, as it demonstrates that the model’s patch partitioning behavior aligns with well-established anatomical context.
\section{Conclusion} 
In this study, we propose the adaptive Morph-Patch Transformer (MPT) to address the challenges of aortic vascular segmentation, specifically the limitations of fixed patch partitioning in traditional Transformer-based models. MPT incorporates two key innovations: a velocity field-guided Morph-Patch strategy that generates morphology-aware patches, and a Semantic Clustering Attention mechanism that aggregates features from semantically similar regions, enabling precise segmentation of fine-grained vessel structures while preserving topological continuity.

Extensive experiments on three public datasets across 2D and 3D settings demonstrate that MPT consistently outperforms existing approaches, achieving state-of-the-art performance in multiple segmentation metrics. The successful adaptation of the Morph-Patch strategy to the Mamba framework further validates its robustness and generalizability in tubular structure segmentation. These results highlight the potential of adaptive, structure-aware patch partitioning for accurate and reliable medical image analysis, with promising implications for clinical cardiovascular diagnostics.
In future work, we aim to further evaluate MPT on a broader range of datasets and anatomical scenarios, paving the way for its practical adoption in clinical applications.
\section{Acknowledgments}
This work was supported by the Key-Area Research and Development Program of Guangdong Province(No. 2025B1111020001), in part by the Natural Science Foundation of Guangdong Province (No. 2023A1515010673), in part by the Shenzhen Science and Technology Innovation Bureau key project (No. JSGG20220831110400001, No. CJGJZD20230724093303007,KJZD20240903101259001), in part by Shenzhen Medical Research Fund (No. D2404001), in part by Shenzhen Engineering Laboratory for Diagnosis \& Treatment Key Technologies of Interventional Surgical Robots (XMHT20220104009), and the Key Laboratory of Biomedical Imaging Science and System, CAS, for the Research platform support.

\bibliography{aaai2026}
\end{document}